\documentclass[a4paper]{article}

\usepackage{INTERSPEECH2019}
\usepackage{multirow}
\usepackage{algpseudocode}
\usepackage{algorithm}
\usepackage{makecell}

\DeclareMathAlphabet{\mathcal}{OMS}{cmsy}{m}{n}

\DeclareMathOperator{\KLD}{KLD}

\DeclareMathOperator{\tf}{tf}

\DeclareMathOperator{\idf}{idf}
\DeclareMathOperator{\tfidf}{tfidf}

\title{Latent Dirichlet Allocation Based Acoustic Data Selection for Automatic Speech Recognition}
\name{Mortaza (Morrie) Doulaty$^{1,*}$, Thomas Hain$^2$}
\address{
  $^1$Microsoft, Germany\\
  $^2$University of Sheffield, UK}
\email{mdoulaty@microsoft.com, t.hain@sheffield.ac.uk}

\newcommand\blfootnote[1]{%
  \begingroup
  \renewcommand\thefootnote{}\footnote{#1}%
  \addtocounter{footnote}{-1}%
  \endgroup
}

\begin{document}

\maketitle
\begin{abstract}
  Selecting in-domain data from a large pool of diverse and out-of-domain data is a non-trivial problem. In most cases simply using all of the available data will lead to sub-optimal and in some cases even worse performance compared to carefully selecting a matching set. This is true even for data-inefficient neural models. Acoustic Latent Dirichlet Allocation (aLDA) is shown to be useful in a variety of speech technology related tasks, including domain adaptation of acoustic models for automatic speech recognition and entity labeling for information retrieval.
In this paper we propose to use aLDA as a data similarity criterion in a data selection framework. Given a large pool of out-of-domain and potentially mismatched data, the task is to select the best-matching training data to a set of representative utterances sampled from a target domain. Our target data consists of around 32 hours of meeting data (both far-field and close-talk) and the pool contains 2k hours of meeting, talks, voice search, dictation, command-and-control, audio books, lectures, generic media and telephony speech data. The proposed technique for training data selection, significantly outperforms random selection, posterior-based selection as well as using all of the available data.
\end{abstract}

\blfootnote{$^{*}$Core part of this work was performed while the author was studying at the University of Sheffield}


\section{Introduction}
Bootstrapping an speech recognition system for a new domain is a common practical problem. A typical scenario is to have some limited in-domain data from a target domain that ASR system is being built for and a pool of out-of-domain data, often containing a diverse set of potentially mismatched data. Using all of the available data is not a good choice in some cases, especially when the pooled data contains a lot of mismatched data to your target domain. There are two main concerns about using all of the available data. Some times the performance is sub-optimal compared to carefully selecting a matching set and in some cases the performance can be even worse \cite{wei2014unsupervised,doulatyis15}. The other concern is the amount of computation needed to train the models. If a comparable or ideally a better model can be trained with a fraction of the available data, then it would be more computationally efficient to train with the smaller set. In these cases data selection becomes a crucial problem. The same problem is applicable for adaptation data selection as well, where the aim is to select data for adapting acoustic model using a limited in-domain dataset. 

In this paper we propose to use acoustic Latent Dirichlet Allocation (aLDA) for matching acoustically similar data to the limited in-domain data from a pool of diverse data. aLDA is already applied for domain discovery \cite{doulaty2015lda} and domain adaptation \cite{doulaty2015ldadnn} in automatic speech recognition as well as media entity recognition, such as show and genre identification in information retrieval systems for media archives \cite{doulaty2016ldabbc,kim2009acoustic,kim2013line}. 

Next section briefly discusses LDA and aLDA. Section \ref{sec:exp-setup} describes the experimental setup and how aLDA data selection technique works, followed by the conclusion in section 4 and references.

\section{Acoustic Latent Dirichlet Allocation}
As shown in our previous works \cite{doulaty2015lda,doulaty2015ldadnn,doulaty2016ldabbc}, aLDA domain posteriors have a unique distribution across different domains that can be used to characterise the acoustic scenery. In this work we make use of aLDA domain posterior features as a basis of acoustic similarity in a data selection problem. The idea is that using acoustically similar data to a target domain for training acoustic models should improve the ASR accuracy on that domain. While using all of the available data which does not necessarily match the target domain could potentially harm the accuracy.

LDA is an unsupervised probabilistic generative model for collections of discrete data. Since speech observations are continuous data, first it needs to be represented by some discrete symbols, here called acoustic words. A GMM with $N$ mixture components is employed for this purpose. The index of Gaussian component with the highest posterior probability is then used to represent each frame with a discrete symbol. Frames of every acoustic document of length $T$, $ \mathbf{d}_i = \{\mathbf{u}_1,...,\mathbf{u}_t,...,\mathbf{u}_T\} $ are represented as:
\begin{equation}
v_t=\underset{n}{\arg\max} \; P(G_n | \mathbf{u}_t), \;\; 1 \le n \le N
\vspace{2mm}
\end{equation}
where $G_n$ is a Gaussian component from a mixture of $N$ components. With this new representation, document $\mathbf{d}_i$ is represented as $\tilde{\mathbf{d}}_i = \{v_1,...,v_t,...,v_T\} $.
For each acoustic word $v_t$ in each acoustic document $\tilde{\mathbf{d}}_i$, term frequency-inverse document frequency (tf-idf) can be computed as:
\begin{equation}
w_t = \tfidf(v_t, \tilde{\mathbf{d}}_i, \tilde{\mathbf{D}}) = \tf(v_t, \tilde{\mathbf{d}}_i) \; \idf(v_t, \tilde{\mathbf{D}})
\vspace{2mm}
\end{equation}
where $\tilde{\mathbf{D}}$ is the set of all acoustic documents represented with acoustic words. With each document now represented with tf-idf scores as $\bar{\mathbf{d}}_i = \{w_1,...,w_t,...,w_T\}$, the LDA models can be trained.

\begin{figure}
	\centering
	\includegraphics[width=6cm]{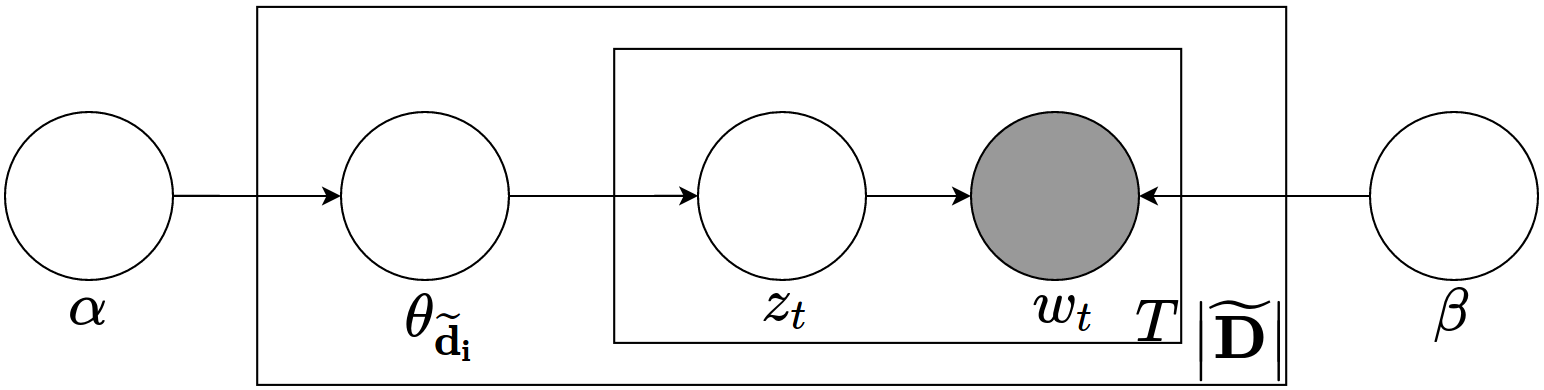}
	\caption{Graphical model representation of LDA}
	\label{fig:lda-graphical-model}
\end{figure}

A graphical representation of the LDA model is shown at Figure \ref{fig:lda-graphical-model}, as a 
three-level hierarchical Bayesian model. In this model, the only observed variables are $w_t$'s. $\alpha$ and $\beta$ are dataset level parameters, $\theta_{\mathbf{\tilde{d}_i}}$ is a document level 
variable and $z_t$ is a latent variable indicating the domain from which $w_t$ was drawn. The following joint distribution is the result of the generative process of LDA:
\vspace{-1mm}
\begin{equation}
p(\theta, \mathbf{z}, \mathbf{\bar{d}} | \alpha, \beta) 
= p(\theta | \alpha) \prod_{t=1}^{T}p(z_t | \theta) p(w_t|z_t,\beta)
\vspace{2mm}
\end{equation}
The posterior distribution of the latent variables given the acoustic document and $\alpha$ and $\beta$ parameters is:
\vspace{-1mm}
\begin{equation}
\label{eq:posterir}
p(\theta, \mathbf{z} | \mathbf{\bar{d}}, \alpha, \beta) = 
\frac{p(\theta, \mathbf{z}, \mathbf{\bar{d}} | \alpha, \beta)}{p(\mathbf{\bar{d}} | \alpha, \beta)}
\vspace{2mm}
\end{equation}
Computing $p(\mathbf{\bar{d}} | \alpha, \beta)$ requires some intractable integrals. A reasonable approximate can be acquired using variational approximation, which is shown to work reasonably well in various applications \cite{blei2003latent}. The approximated posterior distribution is:
\vspace{-1mm}
\begin{equation}
\label{eq:approx-posterior}
q(\theta, \mathbf{z} | \gamma, \phi) = q (\theta | \gamma) \prod_{t=1}^{T}q(z_t | \phi_t)
\vspace{2mm}
\end{equation}
where $\gamma$ is the Dirichlet parameter that determines $\theta$ and $\phi$ is the parameter for the multinomial that generates the latent variables. 

Training minimises the Kullback-Leiber Divergence between the real and the approximated joint probabilities (equations \ref{eq:posterir} and \ref{eq:approx-posterior}) \cite{blei2003latent}: 
\vspace{-1mm}
\begin{equation}
\vspace{2mm}
\underset{\gamma, \phi}{\arg\min} 
\; \KLD \big(
q(\theta, \mathbf{z} | \gamma, \phi)
\; || \; 
p(\theta, \mathbf{z} | \mathbf{\bar{d}}, \alpha, \beta)
\big)
\end{equation}
\vspace{-1mm}

The posterior Dirichlet parameter $\gamma(\mathbf{\bar{d}})$ can be used as features representing the acoustic conditions. These features are used in different tasks, for example for genre and show entity identification and classification tasks \cite{doulaty2016ldabbc, kim2009acoustic, kim2013line, rouvier2009factor} or for domain discovery and adaptation in speech recognition \cite{doulaty2015lda, doulaty2015ldadnn}.

\section{Experimental Setup}
\label{sec:exp-setup}
To evaluate the effectiveness of aLDA for data-selection in ASR, we are trying to solve this practical problem: given a small set of in-domain data and a large pool of out-of-domain and potentially mismatched data, what's the best set of data that can be selected from the pool to train a model for the in-domain data. 

\subsection{Data}
\label{subsec:data}
The in-domain dataset consists of 32 hours of meeting data. Meeting participants used a wide-variety of devices to join the online meetings, including different headsets, earphones with microphones, laptop/table/phone microphone in a far-field setting (arm-length distance) and table-top meeting microphones.
Essentially the data is a mixture of far-field and close-talking in different environments. Table \ref{tab:in-domain-data} summarises some statistics of the in-domain dataset.

\begin{table}[]
\centering
\caption{Statistics of the in-domain dataset}
\label{tab:in-domain-data}
\begin{tabular}{r|l}
Characterisitc  & Notes                                                                                                                                        \\ \hline
Gender          & 37\% female / 63\% male                                                                                                \\ \hline
Nativeness      & 77\% native / 23\% non-native                                                                            \\ \hline
Device          & \makecell[l]{53\% laptop computer \\ 11\% desktop computer \\ 19\% mobile phone \\ 9\% tablet \\ 8\% other devices} \\ \hline
Distance to microphone & 27\% far-field / 73\% close-talk                                      
\end{tabular}
\end{table}

Meetings are mostly real discussions about IT-related topics and there was no control on the participants' recording and environmental conditions. From this in-domain set, 10 hours is used as the dev set and 22 hours as the test set.

The pool of out-of-domain dataset consists of 2000 hours of diverse and multi-domain data. Table \ref{tab:out-of-domain-data} summarises the amount of data for each domain.

\vspace{5mm}
\begin{table}[]
\centering
\caption{Statistics of the out-of-domain dataset}
\label{tab:out-of-domain-data}
\begin{tabular}{l|r|r}
Domain                & \begin{tabular}[c]{@{}r@{}}Duration\\ (hours)\end{tabular} & Percentage\\ \hline
Generic media       & 782    & 39.1\%                                                    \\ \hline
Audio books         & 339    & 17.0\%                                                      \\ \hline
Meeting             & 228    & 11.4\%                                                      \\ \hline
Telephony speech     & 218   & 10.9\%                                                       \\ \hline
Talks               & 172    & 8.6\%                                                       \\ \hline
Command and Control & 112    & 5.6\%                                                       \\ \hline
Lectures            & 111    & 5.6\%                                                       \\ \hline
Dictation           & 38     & 1.9\%                                                       \\ \hline
\textbf{Total}      & \textbf{2000}  & \textbf{100\%}                                           
\end{tabular}
\end{table}

Around 39\% of the pooled data belongs to the generic media domain which includes professional and amateur media recordings from radio, TV, pod-casts and YouTube. Meeting data (which is considered to be the best matching data for our in-domain data) is only 11\% of the pooled data and they were not a part of the in-domain meeting recordings.

The data used for language modelling is fixed in all experiments and includes around 200 million words from Wikipedia, TedTalks, YouTube subtitles and e-books with a vocabulary of size 300 thousand words \cite{heafield2013scalable, Tange2011a}. For the lexicon, a base CMU dictionary was used and for the OOVs, a seq-to-seq g2p model was trained on the base lexicon and used to generate the missing pronunciations \cite{g2pcmu}. 

\subsection{Baseline}
The purpose of this study is to show how aLDA data selection can improve ASR accuracy of a target domain and for that reason all of the model architectures are the same in all of the experiments and the only difference is the amount of data used for training the acoustic models. For acoustic models, TDNN-LSTM model with 3 layers and 1024 nodes in each layer was trained using the lattice-free MMI objective function \cite{povey2016purely} in Kaldi toolkit \cite{povey2011kaldi}. During decoding a pruned 3-gram language model was used to generate lattices and the lattices were then rescored using a 5-gram language model. Table \ref{tab:baseline-results} presents the WER for the test set and for its far-field and close-talk subsets. WER for the far-field subset of the test set is very high and that shows how challenging this dataset is.

\vspace{5mm}
\vspace{10mm}
\begin{table}[]
\centering
\caption{Baseline results}
\label{tab:baseline-results}
\begin{tabular}{c|c|c|c}
\multirow{2}{*}{Model} & \multicolumn{3}{c}{WER}         \\ \cline{2-4} 
                       & Overall & Far-field & Close-talk \\ \hline
Baseline with all data & 29.4    &   53.4    & 20.9      
\end{tabular}
\end{table}

\subsection{aLDA Data Selection}
All of the in-domain data was used for training the aLDA model with the procedure described in section 2 using a vocabulary size of 1024 (number of Gaussian mixture components) and 2048 latent domains. Both these values were selected based on our previous experiments \cite{doulaty2015ldadnn,doulaty2016ldabbc}.
The trained aLDA model was then used to get the posterior Dirichlet parameter $\gamma$ for all of the utterances in the training, dev and test set. The posterior vectors from the dev set were then clustered into 512 clusters using k-means clustering algorithm and the centroid of each cluster was used to represent each cluster. 

An iterative approach was used to select the matching data from the pool of out-of-domain data. For each $\gamma_i$ (centroid of the cluster $i$) the distance to all of the utterances in the training set is computed as:

\begin{equation}
    \Phi(\gamma_i, \gamma_j), \forall \gamma_j \in \mathbf{S^{trn}}
    \vspace{2mm}
\end{equation}
where $\mathbf{S^{trn}}$ is the set of all Dirichlet posterior vectors of the training set and $\Phi$ is the cosine distance between the two vectors defined as:
\vspace{4mm}
\begin{equation}
\Phi(\gamma_i, \gamma_j) = 1 - \frac{\gamma_i  \gamma_j}{||\gamma_i||_{_2}\; ||\gamma_j||_{_2}}
\label{eq:cosdist}
\vspace{2mm}
\end{equation}
The closest utterance (in terms of cosine distance between the Dirichlet posteriors and cluster centroid) that was smaller than a $\lambda$ threshold was added to the selection and was removed from the pool. This iterative process continued until either the minimum distance criterion could not be met for all of the $\gamma_i$ or the pool was depleted. Algorithm 1 shows this iterative process.

Tuning the $\lambda$ threshold requires exploration of a range of values. In our experiments we found that this threshold value is not very sensitive and values in the range of 0.1 to 0.25 resulted in sensible amounts of data. In the final experiment a threshold value of 0.2 was used. This threshold value can also be used to control the amount of data being selected as well if there is budget on the amount of data.

\vspace{5mm}
\begin{algorithm}
\label{algorithm_block}
\caption{Data-selection based on Dirichlet posterior}
\begin{algorithmic}
\State{\textbf{Input:} Training data $\mathbf{S}^{trn}$ of $M$ utterances, \\Training set Dirichlet posteriors $\{\gamma_{1}^{trn},\ldots,\gamma_{M}^{trn}\}$,\\Dev set posterior centroids  $\{\gamma_{1}^{dev},\ldots,\gamma_{N}^{dev}\}$, \\Distance threshold $\lambda$\\}

\State{\textbf{Initialize:} $\mathbf{S}^{new} = \{ \}; \; count=0;$\\}

\While{$\mathbf{S}^{trn} \ne \varnothing $}
 \State{$count = 0$}
 \For{ All $\gamma_{i}^{dev} \in \left\{\gamma_{1}^{dev},\ldots,\gamma_{N}^{dev}\right\}$}
     \State{$d = \min \; \Phi(\gamma_i^{dev}, \gamma_j^{trn})\; \forall \gamma_j^{trn} \in \{\gamma_{1}^{trn},\ldots,\gamma_{M}^{trn}\}$}
     
     \If{$d < \lambda$}
        \State{$j^{*} = \underset{j}{\arg\min} \; \Phi(\gamma_i^{dev}, \gamma_j^{trn})$}
        \State{Remove $\gamma_{j^*}^{trn}$ from $\{\gamma_{1}^{trn},\ldots,\gamma_{M}^{trn}\}$ set\\}
        \State{$\mathbf{S}^{trn} = \mathbf{S}^{trn} \setminus \{ s^{trn}_{j^*} \} $ \\}
        \State{$\mathbf{S}^{new} = \mathbf{S}^{new} \cup \{ s^{trn}_{j^*} \}$ \\}
        \State{$count = count + 1$\\}
     \EndIf
 \EndFor
 \If{$count == 0$}
    \State{break}
 \EndIf
\EndWhile
\\
\State{\textbf{Output:} $\mathbf{S}^{new}$}

\end{algorithmic}
\end{algorithm}

\subsection{Combining Text LDA with aLDA}
Text-based LDA (tLDA) can also be used to further improve the aLDA data selection. The idea is that aLDA captures acoustic similarities in the data and tLDA can further help with linguistic content's similarity. tLDA is already shown to improve classification accuracy in LDA based acoustic information retrieval \cite{doulaty2016ldabbc} as well as language modelling tasks \cite{MGB_ASR_SHEF, deena2016combining, deena2017semi,deena2019recurrent,saz2018lightly}. Training tLDA models followed a similar procedure to aLDA and a comparable number of latent topics and vocabulary size was used.
In our experiments tLDA on its own was not outperforming the baseline and hence those results are not included in this paper. An explanation for it could be the fact that pure linguistic similarity does not necessarily mean that the acoustic conditions are similar as well and thus cannot compensate for the acoustic mismatch.

Different approaches for combining aLDA and tLDA scores were examined. Including but not limited to linear combination of posteriors, two level hierarchical search and two independent search followed by union.  At the end using two approaches independently and then combining the selected data resulted in the best performance.

\subsection{Results and Discussion}
In this section LDA based data selection is compared against random selection, using all of the available data (2000 hours) and phone-posterior based data selection \cite{doulaty2016automatic}. Table \ref{tab:results} summerises the results of the experiments. For the random selection two budgets of 500 and 1000 hours are used and each experiment is repeated 2 times and an average value plus the standard deviation of the runs are provided (due to the data size and computation time this experiment could not be repeated more). Using all of the available data, the WER on the test set is $29.4$. Phone-posterior based selection with a predefined budget of 1000 hours yields a WER of 29.0 which is slightly better than using all of the available data, but savings on computation time is massive (50\% less data used for training the model). aLDA method selects $49.7\%$ of the data and brings down the error by $0.9\%$ absolute. Combing aLDA with tLDA further reduces the error to $28.3\%$ while selecting only 108.5 hours more data.

The results presented in table \ref{tab:results} show the effectiveness of the proposed aLDA data selection and how it can be further improved by using tLDA. 

\vspace{4mm}
\begin{table}[]
\centering
\caption{WER and amount of data for different data selection methods}
\label{tab:results}
\begin{tabular}{l|r|l}
Method                            & \begin{tabular}[c]{@{}r@{}}Amount of data\\ (hours)\end{tabular} & WER              \\ \hline
\multirow{2}{*}{Random selection} & 500                                                              & 31.5 ($\pm$2.00) \\ \cline{2-3} 
                                  & 1000                                                             & 30.1 ($\pm$0.98) \\ \hline
All of data                       & 2000                                                             & 29.4             \\ \hline
Phone-posterior                   & 1000                                                             & 29.0             \\ \hline
aLDA                              & 995.4                                                            & 28.5             \\ \hline
aLDA + tLDA                       & 1103.9                                                           & 28.3            
\end{tabular}
\end{table}

\subsection{Analysis of the Selected Data}
From the pool of 2000 hours, aLDA technique selected 995.4 hours. In this section the selected data is analysed to understand which parts of the training data was found to be the best match to the target in-domain data. 

The training pool consists of data from 8 domains: audio books, command and control, dictation, generic media, lectures, meetings, talks and telephony speech. From these domains only the meeting data seems to be the best match, at least in terms of the domain tags associated with each component. As mentioned in section \ref{subsec:data}, the meeting data in our training set is not a part of the test set recordings, but rather some  generic and diverse meeting data. It includes data from the AMI \cite{carletta2006ami} and ICSI \cite{janin2003icsi} projects as well as some other internal and external sources and in that sense it's not considered as an strictly in-domain data. 

\begin{table}[]
\centering
\caption{Amount of selected data by aLDA}
\label{tab:results}
\begin{tabular}{l|r|r}
Component                            & \begin{tabular}[c]{@{}r@{}}Duration\\ (hours)\end{tabular} 
& \makecell[l]{Percentage\\of domain}  \\ \hline
Generic media                        & 317.5   & 40.6\%                                               \\ \hline
Meeting                              & 205.5   & 90.1\%                                                 \\ \hline
Audio books                          & 147.6   & 43.5\%                                                 \\ \hline
Talks                                & 136.9   & 79.6\%                                                 \\ \hline
Lectures                             & 94.6    & 85.2\%                                                 \\ \hline
Telephony speech                     & 84.3    & 38.7\%                                                \\ \hline
Dictation                            & 6.3     & 16.6\%                                                 \\ \hline
Command and Control                  & 2.7     & 2.4\%                                                 \\ \hline
\textbf{Total} & \textbf{995.4}   & n/a                       \\ 
\end{tabular}
\end{table}

The majority of the selected data belongs to the generic media domain (which was the predominant class in our pool), also almost all of the available meeting data was selected showing that it was a very good match to our in-domain meeting data, at least compared to other data sources. Other interesting observation is the amount of data from dictation and command and control domains, where in total only 8 hours is selected. Checking those data, they are very clean audio. Command and control data set has a lot of very short utterances (single words) and that could contribute to the LDA domains posterior mismatch and not being selected.

In the previous section it was shown that including the data from tLDA selection improves the ASR performence while adding only 108 extra hours. Inspecting those extra data reveals that most of them are selected from talks and telephone speech (35h and 65h respectively). Suggesting that the textual similarities of those domains was picked up by the tLDA and we end up using all of the available talks data in the training of the aLDA+tLDA model. Those extra data improves the accruacy by 0.2\% absolute.

\section{Conclusions}
Selecting matching data to a small set of in-domain data from a large pool of out-of-domain and mismatched data is a non-trivial problem. This problem arises in many practical applications of speech recognition where the task is to build an ASR system for a new target domain where there is a very limited amount of data is available. Often using all of the potentially mismatched data results in sub-optimal and poor performance compared to carefully selecting a matching subset.

In this paper aLDA based data selection is proposed for the first time and its effectiveness is experimented on a large dataset. Our in-domain dataset contains 32 hours of meeting data (mixed far-field and close-talking) and the pool of out-of-domain data consists of 2000 hours of data from very diverse domains. Using all of the available data, the baseline WER is 29.4\%. Using the proposed iterative data selection technique and with slightly less than half of the training data the overal WER on our 20-hour test set is 0.9\% absolute better than using all of the available data. Combining aLDA with tLDA further reduces the WER to 28.3\%.

Future work can include automatic distance threshold finding, exploring the effectiveness of aLDA data selection with data augmentation, finding better ways to combine aLDA and tLDA and further analysis of the selected data by aLDA+tLDA.

\section{Acknowledgements}
The first author would like to thank Trevor Francis for supporting parts of this work.

\vspace{4mm}
\bibliographystyle{IEEEtran}

\bibliography{mybib}


\end{document}